\documentclass[a4paper,conference]{IEEEtran}
\usepackage{mathtools}
\usepackage{stmaryrd}
\usepackage{graphicx}
\usepackage{amsthm,amsmath,amssymb,amsfonts}
\usepackage{booktabs}
\usepackage{wrapfig}
\usepackage{subfig}
\usepackage{placeins}
\usepackage{romannum}
\usepackage{array}
\usepackage{stfloats}
\usepackage{lipsum}
\usepackage{todonotes}

\def\eqref#1{equation~\ref{#1}}

\def\1{\bm{1}}

\DeclareMathAlphabet{\mathsfit}{\encodingdefault}{\sfdefault}{m}{sl}
\SetMathAlphabet{\mathsfit}{bold}{\encodingdefault}{\sfdefault}{bx}{n}

\newcolumntype{P}[1]{>{\centering\arraybackslash}p{#1}}
\ifCLASSINFOpdf
\else
\fi
\def\vec#1{\mathchoice{\mbox{\boldmath  $\displaystyle\bf#1$}}
{\mbox{\boldmath  $\textstyle\bf#1$}}
{\mbox{\boldmath  $\scriptstyle\bf#1$}}
{\mbox{\boldmath  $\scriptscriptstyle\bf#1$}}}

\newcommand{\norm}[1]{\|#1\|} 
\begin{document}
%
\title{Learning Sparse Filters In Deep Convolutional Neural Networks With A $l_1/l_2$ Pseudo-Norm}

\author{\IEEEauthorblockN{Anthony Berthelier \\ and Yongzhe Yan}
\IEEEauthorblockA{Universite Clermont Auvergne\\
Institut Pascal\\
Clermont-Ferrand, France\\
anthony.berthelier@uca.fr\\
yongzhe.yan@uca.fr}
\and
\IEEEauthorblockN{Thierry Chateau\\ and Christophe Blanc}
\IEEEauthorblockA{Universite Clermont Auvergne\\
Institut Pascal\\
Clermont-Ferrand, France\\
thierry.chateau@uca.fr\\
christophe.blanc@uca.fr}
\and
\IEEEauthorblockN{Stefan Duffner\\ and Christophe Garcia}
\IEEEauthorblockA{INSA Lyon\\
LIRIS\\
Lyon, France\\
stefan.duffner@insa-lyon.fr \\
christophe.garcia@insa-lyon.fr }}


%


\maketitle

\begin{abstract}
While deep neural networks (DNNs) have proven to be efficient for numerous tasks, they come at a high memory and computation cost, thus making them impractical on resource-limited devices. However, these networks are known to contain a large number of parameters. 
Recent research has shown that their structure can be more compact without compromising their performance.\\
In this paper, we present a sparsity-inducing regularization term based on the ratio $l_1/l_2$ pseudo-norm defined on the filter coefficients. 
By defining this pseudo-norm appropriately for the different filter kernels, and removing irrelevant filters, the number of kernels in each layer can be drastically reduced leading to very compact Deep Convolutional Neural Networks (DCNN) structures. 
Unlike numerous existing methods, our approach does not require an iterative retraining process and, using this regularization term, directly produces a sparse model during the training process. 
Furthermore, our approach is also much easier and simpler to implement than existing methods.
Experimental results on MNIST and CIFAR-10 show that our approach significantly reduces the number of filters of classical models such as \textit{LeNet} and \textit{VGG} while reaching the same or even better accuracy than the baseline models. 
Moreover, the trade-off between the sparsity and the accuracy is compared to other loss regularization terms based on the $l1$ or $l2$ norm as well as the SSL \cite{10.5555/3157096.3157329}, NISP \cite{DBLP:journals/corr/abs-1711-05908} and GAL \cite{DBLP:conf/cvpr/LinJYZCYHD19} methods and shows that our approach is outperforming them.
\end{abstract}


%
\IEEEpeerreviewmaketitle

\section{Introduction}

Since the advent of \textit{Deep Neural Networks} (DNNs) and especially \textit{Deep Convolutional Neural Networks} (DCNNs) and their massively parallelized implementations \cite{krizhevsky2012imagenet}\cite{lecun2015deep}, deep learning based methods have achieved state-of-the-art performance in numerous visual tasks such as face recognition, semantic segmentation, object classification and detection, etc.  \cite{krizhevsky2012imagenet}\cite{Simonyan2015}\cite{Szegedy2015}\cite{He2015}\cite{He20152}. 
Accompanied with the high performance, also high computation capabilities and large memory resources are needed as these models usually contain millions of parameters. 
These issues prevent them from running on resource-limited devices such as smartphones or embedded devices. 
Network compression is a common approach in this context, i.e.~to reduce the inherent redundancy in the parameters and thus in the computation.\\  

   \begin{figure}[tbp]
      \centering
      \includegraphics[width=\columnwidth]{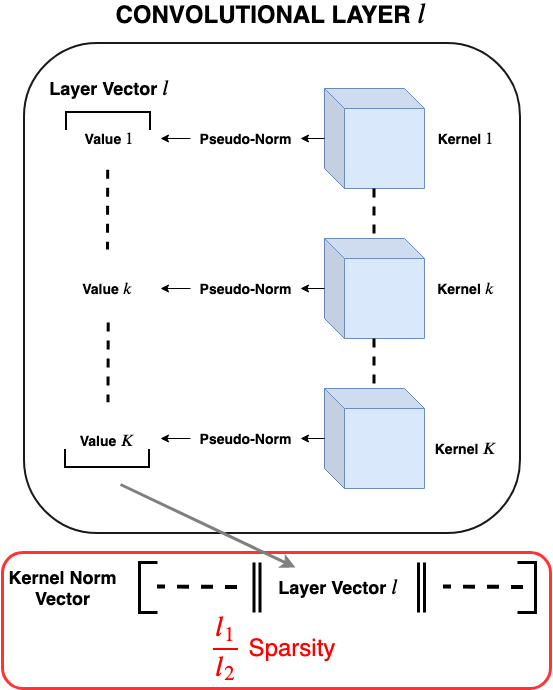}
      \caption{Visual representation of our method and the computation of the kernel norm vector using a pseudo-norm.}
      \label{fig:intro}
   \end{figure}

Numerous methods have been developed to obtain compact DNNs. Since a large number of these networks are built upon convolutional layers and since the convolution operations are the most computationally demanding, we are focusing on the reduction of these layers. 
A simple reduction strategy consists in removing non-relevant filters using pruning methods. For example, Li \textit{et al.} \cite{Li2017} proposed to remove filters that are identified as having a small effect on the output accuracy. Another approach by Luo  \textit{et al.} \cite{Iccv2017} is evaluating information at the filter level using statistical and optimization methods.\\

Our approach is motivated and inspired by (1) previous works demonstrating the redundancy among the weights of a DCNN \cite{speedy}; (2) numerous sparsity methods proposed in the literature~\cite{optispatsity} and (3) the fact that these sparsity methods have rarely been used to remove unimportant weights during training~\cite{10.5555/3157096.3157329}. 
We therefore propose a new strategy, based on $l_1/l_2$-norm, to obtain a subset of kernels with all weights equal to zero (such as that the associated filters can be removed). 
The main idea is to express the filter reduction problem by introducing sparsity on a set of pseudo-norms computed on each kernel but not directly on the kernels actual values.
Figure~\ref{fig:intro} illustrates the general idea of our method. Each kernel of the network is transformed to a single value using a pseudo-norm. 
All these values are concatenated into a global vector (its size being the number of filters) called kernel norm vector. Our global kernel-sparsity is defined by the sparsity on this vector and is estimated by a $l_1/l_2$-norm ratio.  
Since a kernel with all weights equal to zero produces a pseudo-norm of zero, the number of filters can be reduced by enforcing sparsity on the kernel norm vector. 
In this paper, we propose the $l_1/l_2$-norm for two reasons: (1) the so-called $l_1/l_2$-norm is a simple group norm to implement and (2) the use of the $l_1$-norm can increase the performance, interpretability and sparsity of a model \cite{huangsparsity2010} \cite{simulVarSel}\cite{Analysisofvariance} combined with the $l_2$-norm allows to converge to stable solution and maintain sparsity at a good level.\\ 


We propose a $l_1/l_2$-norm computed on the global vector (vector of kernel pseudo-norms) such that adding this sparsity term to minimize to the global loss will reduce the number of (non-zero) filters of a DCNN. 
Compared to other approaches, our method presents several advantages:
\begin{enumerate}
   \item All steps are done during training, i.e.~no additional fine-tuning operations are needed.\\
   \item Our method being based on simple $l_1$ and $l_2$ norms, is straightforward to implement and compute compared to other methods that remove weights during training.\\
   \item As we are keeping track of the evolution of the network at every step during training, it is possible to choose the best model based on a trade-off between compression and accuracy.\\
\end{enumerate}


In the following, we will first present existing work related to network pruning and weight sparsity, in Section \ref{related}. 
In Section \ref{Method}, we describe our $l_1/l_2$ pseudo-norm method. Finally, in Section \ref{experiments}, we show experimental results of our method with \textit{LeNet} and \textit{VGG} network architectures trained on the MNIST and CIFAR-10 datasets. 
We demonstrate that our method is able to significantly improve the sparsity among convolutional layers in these DCNNs without significant drops in accuracy.

\section{Related Work}
\label{related}

 Many studies have been done on DNN compression. Knowledge distillation \cite{Dauphin2013}\cite{Ba2013}  tackles the problem of transferring the encoded information from a bigger model into a smaller one. Lowering numerical precision is also an extensive field \cite{Gupta2015}\cite{Courbariaux2015a}\cite{Williamson1991}. Many works, are focused on designing compressed and optimal models architectures. SqueezeNet \cite{Iandola2016} and MobileNets \cite{Howard2012} both propose structures of convolutional layers to improve memory and computation time. Some Neural Architecture Search (NAS) \cite{Miikkulainen2017}\cite{Tan2018}\cite{DBLP:conf/eccv/HeLLWLH18} methods use reinforcement learning and genetic algorithms to search the best possible networks designs for a given task. Depending on the size of the search space, finding an optimized model with these methods can be enormously time-consuming. However, the most promising approaches try to reduce the model redundancy and among them: parameter quantization \cite{Han2016}\cite{Choi2016} and network pruning \cite{Signorini1995}\cite{Anwar2015}\cite{Molchanov2016a}\cite{Signorini1995}\cite{Li2017}\cite{Iccv2017}. Our method can be classified in this last category. 

\subsection{\textbf{Network Pruning}}
Pruning methods are aiming to remove unimportant parameters of a neural network. Han \textit{et al.} \cite{Han2016}\cite{Signorini1995} proposed to prune parameters of \textit{AlexNet} and \textit{VGG} with connection pruning by setting a threshold and removing any parameters under it. As opposed to our method, most of the reduction is done on fully connected layers and not on convolutional layers. However, compression of convolutional layers is essential nowadays as new DNNs are mostly DCNNs with fewer fully connected layers \textit{e.g.,} only 3.99\% parameters of \textit{Resnet} \cite{He20152}. Closer to our approach, structured pruning methods are removing directly structured parts e.g., kernels or layers, to compress CNNs. Li \textit{et al.} \cite{Li2017} used $l_1$-norm to remove filters. He \textit{et al.} \cite{channelHE} used a LASSO regression based channel selection to prune filters. Channel pruning methods are preferred on widely-used DCNNs. For example, the selection of unimportant feature maps can be done using $l_1$-regularization \cite{Liu2017}.\\
These past few years, numerous networks compression algorithms using pruning methods and achieving state-of-the-art results have emerged. Yu \textit{et al.} \cite{DBLP:journals/corr/abs-1711-05908} proposed a neurons importance score propagation (NISP) method based on the response of the final layers to evaluate the pruning impact of the prior layers. Zhuang \textit{et al.} \cite{NIPS2018_7367} developed discrimination-aware losses in order to determine the most useful channels in intermediate layers. Some methods such as Filter Pruning Via Geometric Median (FPGM) \cite{he2019filter} are not focused on pruning filters with less importance but only by evaluating their redundancy. Similarly, Lin \textit{et al.} \cite{DBLP:conf/cvpr/LinJYZCYHD19} tackled the problem of redundant structures by  proposing a generative adversarial learning method (GAL) (not only to remove filters, but also branches and blocks). \\
Still, standard pruning methods usually construct non structured and irregular connectivity in a network, leading to irregular memory access. In most of these approaches, the DNN is trained first. Then each parameter is evaluated to understand if it brings information to the network. If not, the parameter is removed. Therefore, a fine-tuning needs to be performed afterwards to restore the model accuracy. These steps take time. Most of them are done offline and need costly reiterations of decomposing and fine-tuning to find an optimal weight approximation maintaining high accuracy and high compression rate. Unlike these methods, our approach is able to directly increase the sparsity of the network during training, identifying which kernels to prune without any considerable extra computational overhead.\\

\subsection{\textbf{Weight Sparsity}}
An important factor for the compression of a model is its sparsity \textit{i.e.} the number of parameters set to zero. However, this sparsity must be structured in order to be memory-efficient and time-efficient. Liu \textit{et al.} \cite{7298681} obtained a sparsity of 90\% on \textit{AlexNet} with only 2\% accuracy loss using sparse decomposition and a sparse matrix multiplication algorithm. This method also employed group Lasso \cite{modelselectionlasso}, an efficient regularization to learn sparse structures. It is also used by Wen \textit{et al.} \cite{10.5555/3157096.3157329} to regularize the structure of a DNN at different levels (\textit{i.e.} filters, channels, filter shapes and layer depth). This approach leads to DNNs with reduced computational cost and efficient acceleration due to the structured sparsity induced by the method. 
We propose to use a different type of regularization based on the norm ratio $l_1/l_q$ \cite{efficientl1lq}\cite{optispatsity}. 
It
allows to dynamically maximize the sparsity of a model with one hyper-parameter ($q$) without additional iterations and severe drops in accuracy while being straightforward to implement.

\section{Training with kernel-sparsity}
\label{Method}
We mainly focus on inducing sparsity on convolutional layers to regularize and compress the structure of DCNNs during the training steps. We propose a generic method to regularize DCNNs using the $l_1/l_2$ pseudo-norm.\\ 


   \begin{figure*}[t!]
      \centering
      \includegraphics[width=1\textwidth]{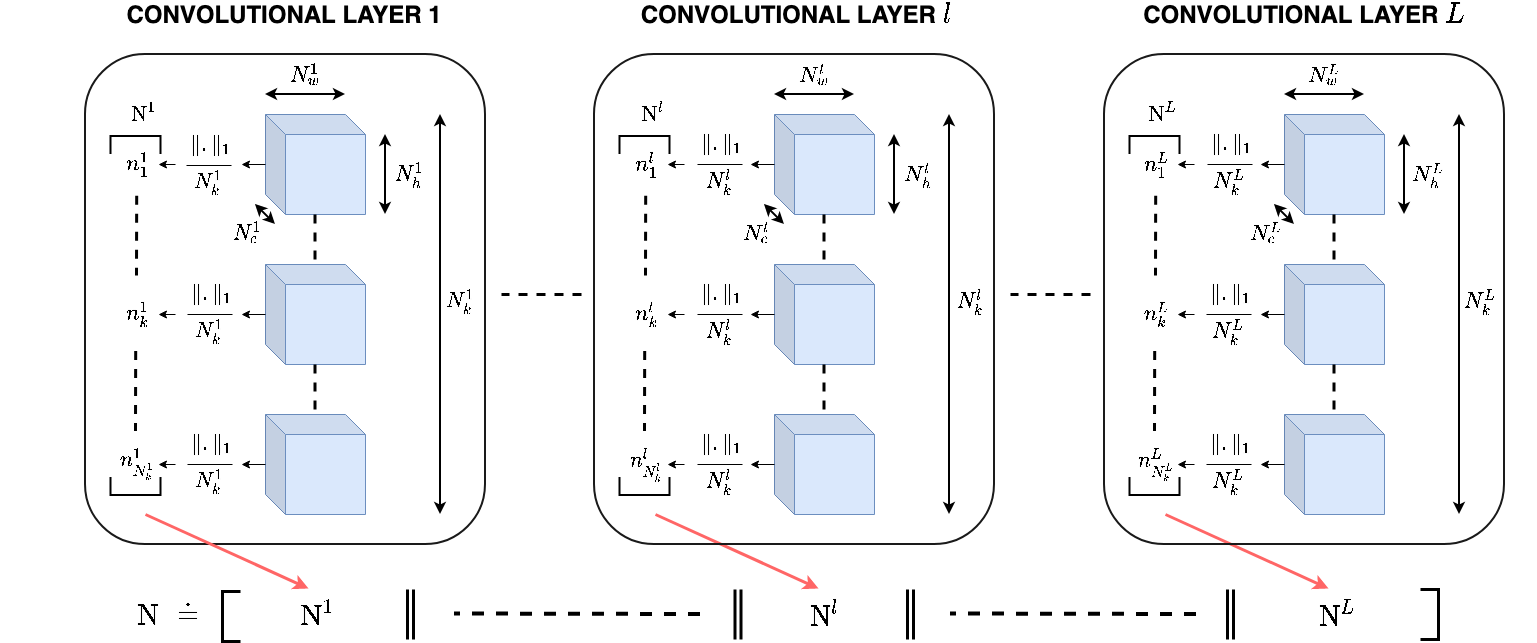}
      \caption{Visualization of the computation of the kernels pseudo-norm and how the global kernel norm vector $\vec{N}$ is obtained.}
      \label{fig:resume}
   \end{figure*}

\subsection{\textbf{Kernel-sparsity regularization}}

Let ${\cal N}$ be a DCNN with $L$ convolutional layers. We define $W^{l,k}$ as the  $k^{th}\in\{1,..N_k^l\}$ 3d-tensor (kernel) associated with the $l^{th}$ convolutional layer. Thus, a weight of kernel $k$ in the convolutional layer $l$ is defined as:   
and $W_{w,h,c}^{l,k}\in \mathbb{R}^{N_w^l,N_h^l,N_c^l}$ the (width, height, channel) weight of kernel $k$ of layer $l$.

\begin{equation}
W_{w,h,c}^{l,k}\in \mathbb{R}^{N_w^l,N_h^l,N_c^l}
\end{equation}

Here, $w\in\{1,..N_w^l\}$ is the column, $h\in\{1,..N_h^l\}$ is the row and $c\in\{1,..N_c^l\}$ is the channel index of the $k^{th}$ kernel matrix in the convolutional layer $l$. 
The key idea is to express sparsity on pseudo-norms of kernels. Let $n^l_k$ be the pseudo-norm defined by the $l_1$-norm of the flattened kernel $W^{l,k}$: 
\begin{equation}
    n^l_k \doteq \sum_{w=1}^{N_w^l} \sum_{h=1}^{N_h^l} \sum_{c=1}^{N_c^l} \frac{|W_{w,h,c}^{l,k}|}{N_k^l}
\end{equation}
The vector $\vec{N}^l$ concatenates, for layer $l$, the $N_k^l$ norms $n^l_k$:

\begin{equation}
    \vec{N}^l \doteq \bigparallel_{k=1}^{N_k^l} n^l_k
\end{equation}

We introduce kernel-sparsity for a layer as a value linked to the number of kernels of this layer with all weights equal to zeros. Therefore, the kernel-sparsity of layer $l$ can be linked with the number of values of the vector $\vec{N}^l$ equal to zero. Global kernel-sparsity can be expressed from the concatenation of vectors  $\vec{N}^l$ for each layer:

\begin{equation}
    \vec{N} \doteq \bigparallel_{l=1}^{L} \vec{N}^l
\end{equation}

For better understanding, we visualize these operations in Figure \ref{fig:resume}. In order to normalize the value of N, each of its component is divided by the number of values (or norms) that it contains. Finally, the global kernel-sparsity is defined by the sparsity of $\vec{N}$ and can be estimated by a ${l_1}/{l_2}$ ratio function:
\begin{equation}
    {\cal L}_{s}\doteq \cfrac{\norm{\vec{N}}_1}{\norm{\vec{N}}_2} \; .
\end{equation}
Minimizing this term will encourage zero-valued coefficients (numerator), corresponding to the different kernels, while keeping the remaining coefficients at large values (denominator), thus producing convolution layers with few non-zero kernels.

\subsection{\textbf{Training with kernel-sparsity regularization}}

Let ${\cal L}_{\cal N}$ be the 
loss function that is minimized to find the optimal weight configuration for a given task (e.g.~cross entropy).
We propose to simply add the kernel-sparsity regularization term weighted by the coefficient $\lambda \in \mathbb{R}$: 
\begin{equation}
{\cal L}_{all} = {\cal L}_{\cal N} + \lambda {\cal L}_s     \; .
\end{equation}
We will discuss how to set an appropriate values of $\lambda$ in the experimental section.

\subsection{\textbf{Setting kernels to zero}}

Our method induces sparsity in a DCNN, i.e.~the pseudo-norm regularization pushes some kernels to have only zero-valued coefficients. 
However, in practice, during optimization, the actual values of these kernels will not be exactly zero but very small. 
Thus, to compress the network effectively, our approach identifies these kernels during training and forces them to be zero in order to remove them.

More specifically, the algorithm works as follows: each pseudo-norm of the kernels is contained in the global kernel pseudo-norm vector $\vec{N}$. Thus, at each epoch, we normalize the values of $\vec{N}$ so that $\sum_{i=1}^{K} \vec{N}_i = 1$. Sorting these vectors in ascending order will allow us to objectively determine which pseudo-norms are the smallest. 
We then define a percentage (or a threshold) under which the cumulative sum of these sorted values is judged too small to be kept, i.e.~the corresponding filters are considered unimportant and set to zero. 
Once the weights of a kernel are set to zero, they are keeping this value until the end of the training, and these parameters are no longer updated. 
This ensures that the potential errors and imprecision introduced by removing these kernels can be compensated by the remaining kernels during the training converging to a stable solution with high accuracy.


To summarize, our approach consists of two steps at each epoch:
\begin{enumerate}
   \item The $l_1/l_2$ pseudo-norm is computed on each kernel of the model and is integrated to the loss function. Thus the training stage is minimizing the loss function and inducing sparsity at the kernel level, pushing some weights to have a near zero value.\\
  
   \item Sort kernels according to their ascending  normalized pseudo-norm and compute a cumulative sum vector from the sorted normalized pseudo-norm vector. The kernels participating to the cumulative sum under a threshold $t$ are removed. This set of operations aims at keeping kernels that produce more than $t\%$ of the global norm.
   
\end{enumerate}

\section{Experiments}
\label{experiments}
We evaluated the performance of the $l_1/l_2$ pseudo-norm on two classification models (\textit{LeNet} and \textit{VGG}) and two datasets: MNIST and CIFAR-10. Our method is implemented in \textit{Pytorch}, running on various Nvidia GPUs using CUDA. The weights of the networks are initialized randomly and hyper-parameters are selected manually for optimal results. The chosen $\lambda$ value is the one allowing the model to have an accuracy close to its baseline accuracy while sparsifying the most the kernels norms. In all the experiments, the threshold under which the kernels are removed by evaluating the cumulative sum of the smallest norms is set to $1\%$. We found that this value was the best trade-off between a converging accuracy of the models and a slow removal of the kernels during the training phase.

   \begin{figure*}[t!]
      \centering
      \includegraphics[width=0.95\textwidth]{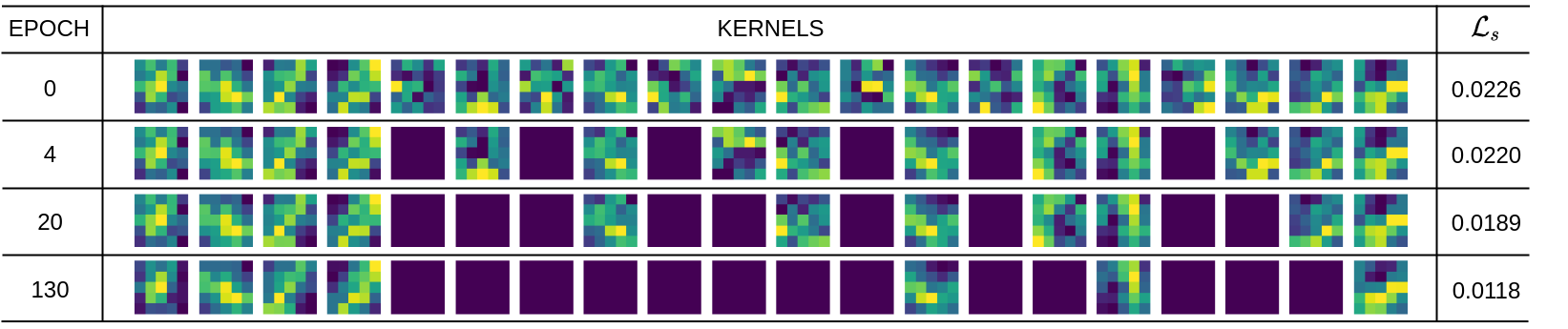}
      \caption{Evolution of the kernels and the global kernel-sparsity regularization term during training. Evaluations are done on the first convolutional layer of \textit{LeNet} on the MNIST dataset.}
      \label{fig:kernelevo}
   \end{figure*}
   
    \begin{figure}[t!]
      \centering
      \includegraphics[width=0.45\textwidth]{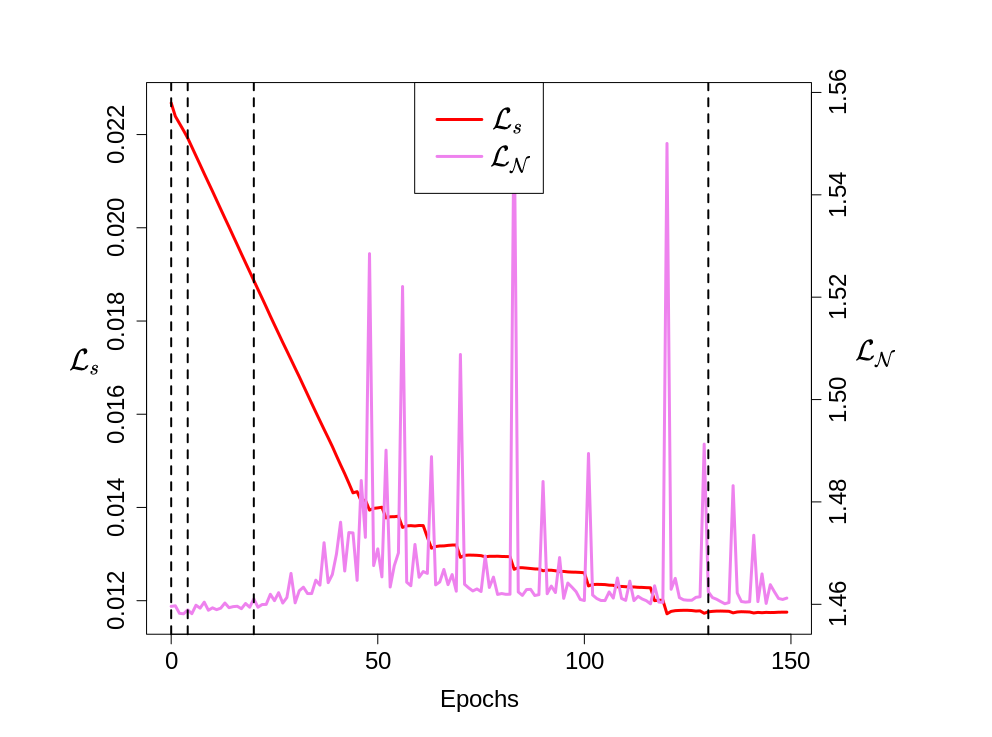}
      \caption{Evolution of the cross-entropy loss function ${\cal L}_{\cal N}$ and the global kernel-sparsity regularization term ${\cal L}_s$ during training with $\lambda=0.5$. Evaluations are done on \textit{LeNet} on the MNIST dataset. Vertical lines show which epochs were taken to construct Figure \ref{fig:kernelevo}.}
      \label{fig:graphloss}
   \end{figure}

\subsection{\textbf{\textit{Experiments on LeNet}}}

In the experiments with \textit{LeNet} \cite{lecun1998gradient}, we investigate the effectiveness of the $l_1/l_2$ pseudo-norm on the MNIST and CIFAR-10 datasets. In order to compare our results with state-of-the-art methods such as SSL \cite{10.5555/3157096.3157329}, NISP \cite{DBLP:journals/corr/abs-1711-05908} and GAL \cite{DBLP:conf/cvpr/LinJYZCYHD19}, we decide to chose the \textit{LeNet} model implemented by \textit{Caffe}. All these methods are using evaluation and regression at different level i.e Lasso-group regression at different level of a convolutional layer in the SSL method, which makes it more complex to implement than our method. There is no data augmentation for the training on both datasets.\\


\textbf{\textit{LeNet on MNIST}}: As previously described, the $l_1/l_2$ pseudo-norm is applied on the filters of a DCNN to penalize them. Hence our method is inducing sparsity among the filters of the convolutional layers in \textit{LeNet}. To visualize the effect on our approach on the kernels, Figure \ref{fig:kernelevo} shows the evolution of the kernels of the first convolutional layer of \textit{LeNet} during a training on MNIST with our kernel-sparsity regularization. We see that the kernel-sparsity term ${\cal L}_s$ is decreasing epoch after epoch and that the number of filters in the layer is also decreasing with it. The complete evolution of the kernel-sparsity ${\cal L}_s$ can be seen on Figure \ref{fig:graphloss}. ${\cal L}_s$ being computed on the pseudo-norm of the kernels and kernels weights being set to zero over time: this result shows the effectiveness of our method.\\

Table \ref{tab:compMNIST} summarizes the results on MNIST of different methods. 
In the best case scenario that we have tested, the $l_1/l_2$ pseudo-norm with a $\lambda$ value set to $0.5$ is able to achieve an accuracy better than the baseline by $0.2\%$. Furthermore the number of filters is dropping drastically in both convolutional layers respectively from 20 to 5 and from 50 to 18. Compared to the other state-of-the-art methods, the $l_1/l_2$ pseudo-norm is able to achieve a better accuracy while penalizing more filters too. We also compare our method to the $l_1$-norm and $l_2$-norm. During our evaluations, both of these norms were able to reach a higher level of sparsity by setting to zero more kernels. However they were never able to reach the same or a better level of accuracy than the baseline.\\

   \begin{figure*}[t!]
      \centering
      \includegraphics[width=1\textwidth]{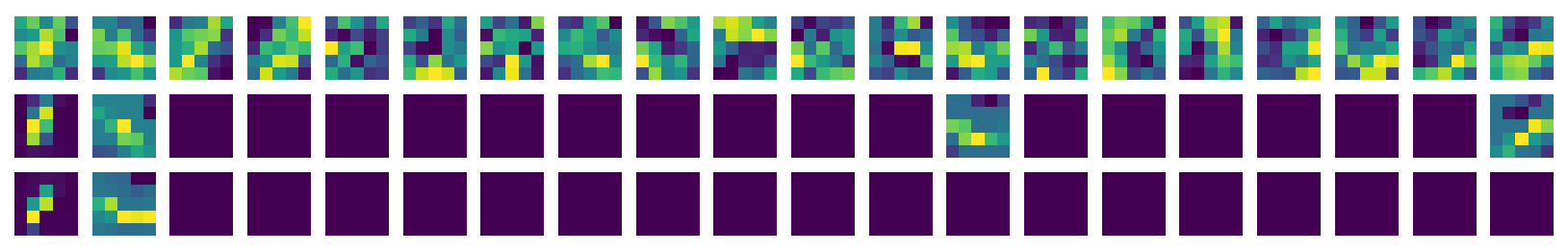}
      \caption{Learned filters of the first convolutional of \textit{LeNet} on MNIST. Top is \textit{LeNet} baseline, middle and bottom are $l_1/l_2$-norm with $\lambda=0.5$ and different level of sparsity. }
      \label{fig:kernelconv1}
   \end{figure*}

\begin{table}[!t]
  \centering
  \small
  \begin{tabular}{P{1cm}P{0.5cm}P{0.8cm}P{1.5cm}P{1.5cm}P{1cm}}
    \toprule
    Method & $\lambda$ & Error & Conv1 Filter \# (Sparsity)& Conv2 filter \# (Sparsity)& Total Sparsity\\
    \midrule
    Baseline & - & \textbf{0.9\%} & 20 & 50 & \textbf{0\%}\\
    \midrule
    $l_1$ & 0.5 & 1.2\% & \hfil4 \newline \null\hfil (80\%)& \hfil5\newline \null\hfil (90\%)& 87.1\%\\
    \midrule
    $l_2$ & 0.5 & 1.2\% & \hfil3 \newline \null\hfil (85\%)& \hfil5\newline \null\hfil (90\%)& 88.6\%\\
    \midrule
    SSL 1 & - & \textbf{0.8\%} & \hfil5 \newline \null\hfil (75\%)& \hfil19\newline \null\hfil (62\%)& \textbf{65.7\%}\\
    \midrule
    SSL 2 & - & 1.0\% & \hfil3 \newline \null\hfil (85\%)& \hfil12\newline \null\hfil (76\%)& 78.6\%\\
    \midrule
    NISP & - & \textbf{0.8\%} & \hfil10 \newline \null\hfil (50\%)& \hfil25\newline \null\hfil (50\%)& \textbf{50.0\%}\\
    \midrule
    GAL & - & 1.0\% & \hfil2 \newline \null\hfil (90\%)& \hfil15\newline \null\hfil (70\%)& 75.7\%\\
    \midrule
    $l_1/l_2$ & 0.5 & \textbf{0.7\%} & \hfil5 \newline \null\hfil (75\%)& \hfil18\newline \null\hfil (64\%)& \textbf{67.1\%}\\
    \bottomrule
  \end{tabular}
  \normalsize
  \caption{Results after penalizing unimportant filters in \textit{LeNet} on MNIST. Baseline is the simple \textit{LeNet} Caffe model. $l_1$ and $l_2$ are the best results found by using the $l_1$-norm and $l_2$-norm regularization on the kernels. SSL, NISP and GAL are the pruning methods respectively from \cite{10.5555/3157096.3157329}, \cite{DBLP:journals/corr/abs-1711-05908} and \cite{DBLP:conf/cvpr/LinJYZCYHD19}. $l_1/l_2$ is our method with $\lambda=0.5$.}
  \label{tab:compMNIST}
\end{table}

To visualize the effect of our method on the parameters, we show the learned filters of the first convolutional layer in Figure \ref{fig:kernelconv1}. For $\lambda=0.5$ and for different level of sparsity, it can be seen that the number of remaining filters can be set to only 2 or 4. Furthermore between the baseline and our method, the accuracy is the same or is increased. This shows that there is effectively a large amount of redundancy between filters and that most of them are not required. Moreover, compared to the baseline, it seems that the remaining filters are more structured, with more regular patterns. This assumption seems especially true when only two filters are remaining. Thus, we arrive at the same conclusion than \cite{10.5555/3157096.3157329}: the baseline has a high freedom in the parameter space and our method is able to obtain the same accuracy by optimizing the filters into more regularized patterns.\\

\textbf{\textit{LeNet on CIFAR-10}}: In order to test the $l_1/l_2$ pseudo-norm and visualize its effect on a more difficult classification task than MNIST, we decided to use the CIFAR-10 dataset with the same \textit{LeNet} model. The results are summarized in Table \ref{tab:compcifar}. The baseline \textit{LeNet} is not performing as well on CIFAR-10 than it is performing on MNIST, i.e.~the classification accuracy is only around 70\%. As a result, the accuracy of our approach also drops but the $l_1/l_2$ pseudo-norm is still able to perform well on this model, even for this classification task. With a $\lambda$ value set to $0.7$, we are able to decrease the number of filters in the first and the second convolutional layers respectively from 20 to 10 and from 50 to 25, which means that half of the filters of \textit{LeNet} are removed. With this configuration, our method performs $1.7\%$ worse than the baseline. We were able to remove up to 80\% of the filters in our experiments, but the resulting accuracy was too low to be interesting (more than 20\% behind the baseline). Hence, more filters are needed in order to classify correctly the CIFAR-10 dataset compared to the MNIST dataset. The best trade-off between filters and accuracy that we found was still with a value of $\lambda$ set to $0.7$. In both convolutional layers, the number of filters is dropping respectively from 20 to 14 and from 50 to 30. This means that our method is able to zero out more than a third of the filters with only a drop of 0.9\% in accuracy. Compared to the $l_1$-norm and the $l_2$-norm, the $l_1/l_2$ pseudo-norm also shows good results. Indeed, both the $l_1$-norm and $l_2$-norm where unable to reach the same level of accuracy and setting to zero as many filters as the $l_1/l_2$ pseudo-norm can do.\\

\begin{figure*}[t!]
      \centering
      \includegraphics[width=1\textwidth]{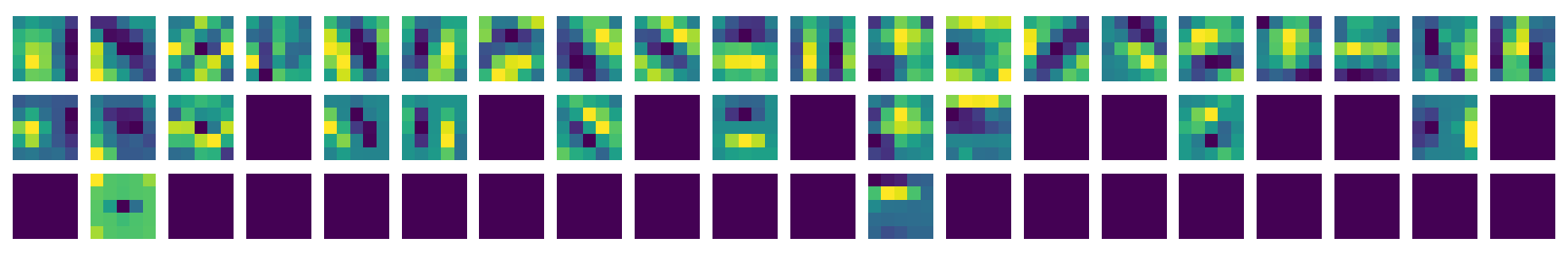}
      \caption{Learned filters of the first convolutional layer of \textit{LeNet} on CIFAR-10. Top is \textit{LeNet} baseline, middle and bottom are $l_1/l_2$-norm with $\lambda=0.7$ and different level of sparsity.}
      \label{fig:kernelconv1cifar}
   \end{figure*}

As previously done with the MNIST dataset, we visualize the learned filters of the first convolutional layer in Figure~\ref{fig:kernelconv1cifar}. From this visualization, we can draw the same conclusion than with the MNIST dataset. The more we are removing filters, the more the remaining ones seems to have a defined structure, as opposed to the baseline where each of the filters seems blurry. It is even more remarkable when we let our algorithm run until only a couple of filters are remaining. Even if the model does not reach a satisfactory accuracy, the two remaining filters have learned remarkable patterns. Thus, the $l_1/l_2$ pseudo-norm is still able to smooth a high freedom of parameter space into fewer filters with more regularized patterns.

\begin{table}[!t]
  \centering
  \small
  \begin{tabular}{P{1cm}P{0.6cm}P{0.8cm}P{1.5cm}P{1.5cm}P{1cm}}
    \toprule
    Method & $\lambda$ & Error & Conv1 Filter \# (Sparsity)& Conv2 filter \# (Sparsity)& Total Sparsity\\
    \midrule
    Baseline & - & \textbf{28.4\%} & 20 & 50& \textbf{0\%}\\
    \midrule
    $l_1$ & 0.7 & \textbf{35.4\%} & \hfil7\newline \null\hfil (65\%)& \hfil14\newline \null\hfil(72\%)& \textbf{70.0\%}\\
    \midrule 
    $l_2$ & 0.7 & 29.8\% & \hfil12\newline \null\hfil(40\%)& \hfil24\newline \null\hfil(52\%)& 48.6\%\\
    \midrule
    $l_1/l_2$ & 0.7 & 30.1\% & \hfil10\newline(50\%)& \hfil25\newline \null\hfil(50\%)& 50.0\%\\
    \midrule
    $l_1/l_2$ & 0.7 & \textbf{29.3\%} & \hfil14\newline \null\hfil(30\%)& \hfil30\newline \null\hfil(40\%)& \textbf{37.1\%}\\
    \bottomrule
  \end{tabular}
  \normalsize
  \caption{Results after penalizing unimportant filters in \textit{LeNet} on CIFAR-10. Baseline is the simple \textit{LeNet} Caffe model. $l_1/l_2$ is our method with different coefficient of regulation $\lambda=0.7$.}
  \label{tab:compcifar}
\end{table}

\subsection{\textbf{\textit{VGG on CIFAR10}}}

To demonstrate the generalization of our method on larger DNNs, we evaluate the performance of our method on the well-known \textit{VGG}~\cite{Simonyan2015}, a deeper model than \textit{LeNet}, with several convolutional layers. A \textit{VGG} model can have different sizes, notably depending on the number of layers. We chose the \textit{VGG11} model with a total of 8 convolutional layers. We implemented it using \textit{Pytorch}, running on various Nvidia GPUs using CUDA. 
The model is trained without data augmentation and evaluated on the CIFAR-10 dataset. In this experiment, the kernels pseudo-norms are not normalized on the full network, which explains why the $\lambda$ values are smaller than the ones used with \textit{LeNet}.\\

With \textit{LeNet}, the $l_1/l_2$ pseudo-norm method was applied on only 2 convolutional layers, with 50 filters at most in the second convolutional layer. In \textit{VGG11} our method is applied on 8 different convolutional layers with a number of filters set to 64 in the first convolutional layer and a maximum of 512 filters in the last four convolutional layers. 
The results are shown in Table \ref{tab:compvgg}. The baseline model, with all the filters and a classical loss function (cross-entropy), obtains an error of 17.6\% on the test dataset. Using the $l_1/l_2$ pseudo-norm with a $\lambda$ set to $0.005$, the model achieves a classification accuracy roughly 1\% inferior to the baseline. However the number of filters is vastly reduced. Moreover, it seems that the deeper we go in the network, the more the proportion of filter sets to zero is important. For example, the second convolutional layer has around 10\% of its filters set to zero while the last convolutional layer has over 65\% of filters set to zero. Thus we could deduce that the last convolutional layers keep less important information for the model than the first ones or that there is more redundancy in the last layers. However, the first convolutional layer seems to be an exception as approximately half of its filters can be removed. We suppose that the shapes learned in the first layer are not decisive for the model and can be balanced by the following layers and the more defined shapes that they have assimilated.\\

\begin{table*}[!t]
  \centering
  \small
  \begin{tabular}{P{1cm}P{0.5cm}P{0.5cm}P{1.25cm}P{1.25cm}P{1.25cm}P{1.25cm}P{1.25cm}P{1.25cm}P{1.25cm}P{1.25cm}P{1cm}}
    \toprule
    Method & $\lambda$ & Error & Conv1 Filter \# (Sparsity) & Conv2 filter \# (Sparsity) & Conv3 Filter \# (Sparsity) & Conv4 Filter \# (Sparsity) & Conv5 Filter \# (Sparsity) & Conv6 Filter \# (Sparsity) & Conv7 Filter \# (Sparsity) & Conv8 Filter \# (Sparsity)& Total Sparsity\\
    \midrule
    Baseline & - & \textbf{17.6\%}& 64 & 128 & 256 & 256 & 512 & 512 & 512 & 512& \textbf{0\%}\\
    \midrule
    $l_1$ & 0.0001 & 19.3\% & 52 (18.3\%) & 128 (0.0\%) & 255 (0.4\%) & 256 (0.0\%)& 175 (65.8\%) & 147 (71.3\%)& 97 (81.1\%) & 123 (75.9\%)& 55.2\%\\
    \midrule
    $l_2$ & 0.005 & \textbf{18.2\%} & 64\newline(0.0\%)& 128 (0.0\%)& 256 (0.0\%)& 256 (0.0\%)& 511 (0.2\%)& 474 (7.4\%)& 434 (15.2\%)& 299 (41.6\%)& \textbf{12\%}\\
    \midrule
    $l_1/l_2$ & 0.005 & 18.8\% & 35 (45.3\%) & 115 (10.2\%) & 238 (7.0\%) & 176 (31.3\%) & 354 (30.9\%) & 195 (61.9\%)& 190 (62.9\%) & 175 (65.8\%)& 46.3\%\\
    \midrule
    $l_1/l_2$ & 0.001 & \textbf{16.8\%} & 64\newline(0.0\%) & 128 (0.0\%) & 256 (0.0\%)& 256 (0.0\%)& 512 (0.0\%)& 512 (0.0\%) & 510 (0.4\%)& 380 (25.8\%)& \textbf{5\%}\\
    \bottomrule 
  \end{tabular}
  \normalsize
  \caption{Results after penalizing unimportant filters in \textit{VGG11} on CIFAR-10. Baseline is the \textit{VGG11} network baseline. $l_1/l_2$ is our method with different coefficient of regulation $\lambda$.}
  \label{tab:compvgg}
\end{table*}

\begin{figure*}[!b]
\centering
\subfloat[First convolutional layer of \textit{VGG11}]{\includegraphics[width=1\columnwidth]{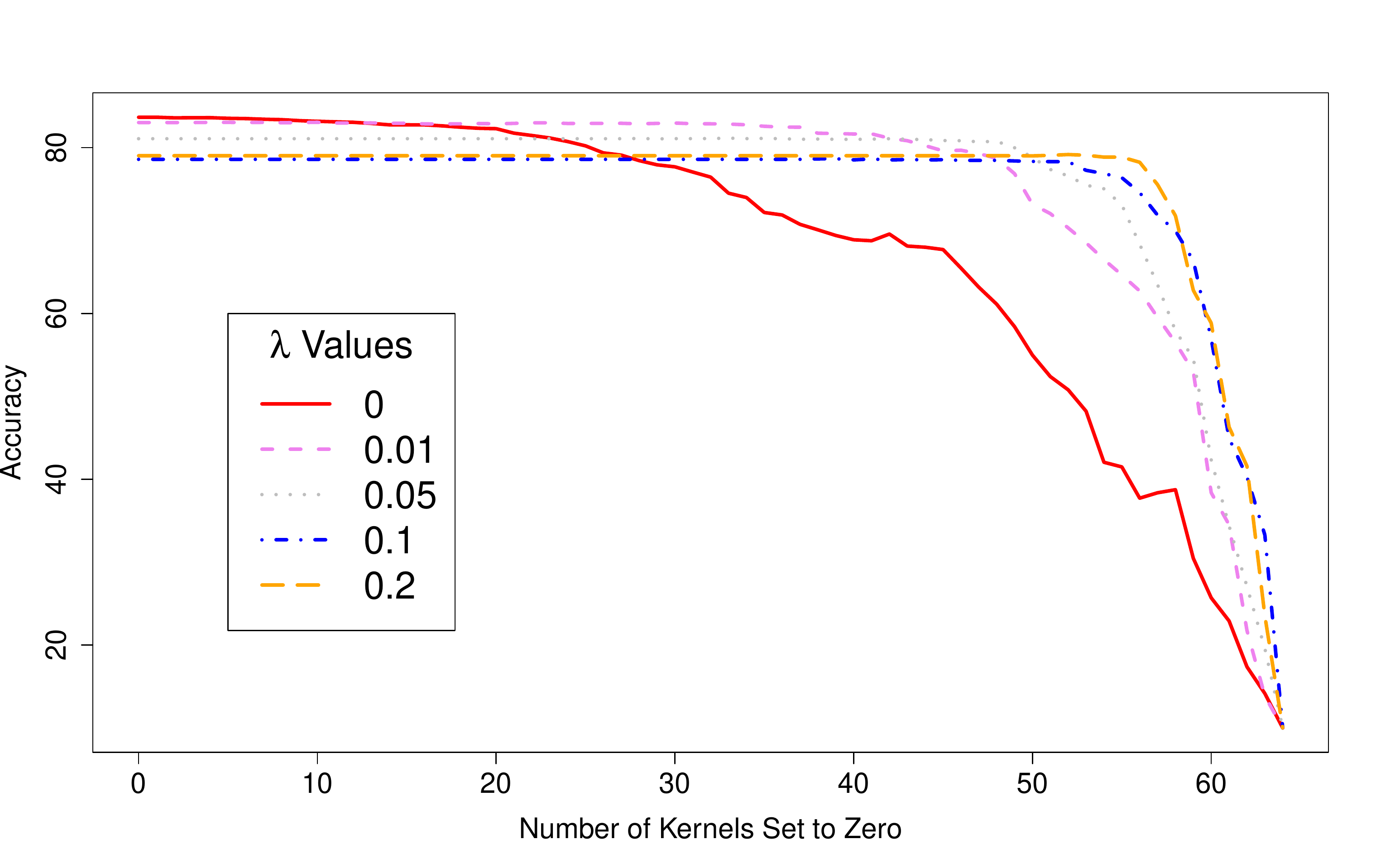}}
\subfloat[Second convolutional layer of \textit{VGG11}]{\includegraphics[width=1\columnwidth]{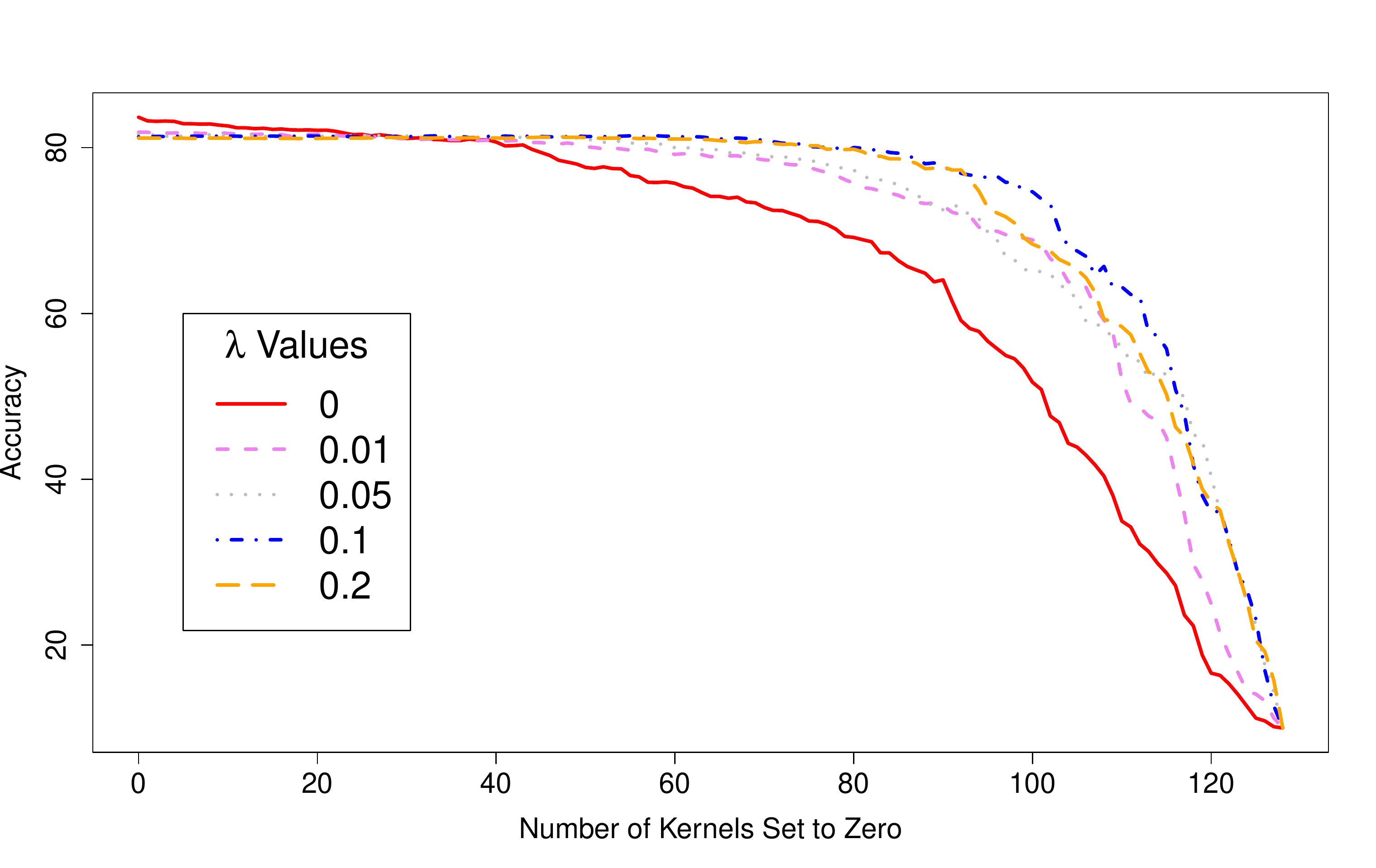}}
\caption{Visualization of the effect of setting kernels to zero in the first two convolutional layers of \textit{VGG11} against the accuracy of the network. Each line represents a different value for the coefficient of regulation $\lambda$ of the $l_1/l_2$ pseudo-norm method.}
\vspace{-4mm}
\label{fig:accukernel} 
\end{figure*} 

By decreasing the $\lambda$ coefficient to $0.001$, we confirm the results that the last convolutional layers seem to contain more filters with non decisive or redundant information than the first ones. Indeed, only the last two layers have filters set to zero. But more importantly, the removal of a few filters in the last two convolutional layers leads to a classification error of only 16.8\%, which is 0.8\% less than the baseline. Thus, our method, by only removing a few filters, is able to achieve a better accuracy than the baseline model.
Compared to the $l_1$-norm and the $l_2$-norm, the $l_1/l_2$ pseudo-norm is also performing well. The $l_1$-norm is able to zero out numerous filters but is unable to achieve a correct level of accuracy, always performing worse than the baseline or our approach. Nearly the same conclusions can be drawn from the $l_2$-norm. Under certain conditions, the $l_2$-norm is able to zero out slightly more filters than our method in the last convolution layers. However, the models are not able to obtain a satisfying accuracy, always around 1\% behind the baseline.\\

In order to conclude this study, we visualize in Figure \ref{fig:accukernel} the evolution of the accuracy of the model against the number of kernel set to zero in the first and second convolutional layers for different coefficient of regularization $\lambda$. When $\lambda=0$, the $l_1/l_2$ pseudo-norm is not taken into account, resulting into the baseline model. The order that the filters are set to zero is determined by the filters pseudo-norm arranged by ascending order. These tests are done at a single convolutional layer level. Meaning that during training, the only filters that are evaluated and set to zero are the ones belonging to the studied layer. The other layers are remaining untouched. We visualize that for both layers, we are able to set numerous filters to zero without a noticeable decrease of the accuracy, even when our method is not active. This result shows that there is unimportant information in the layer and that it is possible to remove it, even if there are no methods that are defined to emphasize this phenomenon. With the implementation of the $l_1/l_2$ pseudo-norm ($\lambda > 0$), we see that (1) more kernels are set to zero before the beginning of the accuracy drop compared to the baseline and (2) a greater $\lambda$ value means that more kernels are zeroed-out but at the price of an inferior accuracy. Based on these conclusions, our method is increasing the sparsity of the filters within a layer, shifting information between them in order to centralize the information. However this sparsity has its limits. The more we force it (with a significant $\lambda$ value), the more we increase the chances to lose important information that could be never recovered.

\section{Conclusion}
\label{conclusion}
In this work, we proposed a new regularization approach for inducing kernel sparsity in DCNNs based on the $l_1/l_2$ pseudo-norm. This method reorganizes the weights of the convolutional layers in order to learn more compact structures during training. These compact DCNNs can reach almost the same accuracy as the original models and in some cases even perform better. Our experiments have demonstrated the benefits of our approach and its generalization to deep structures: it is straightforward to implement, it operates during the training process and it is possible to choose between the compactness and the accuracy of the model. So far, we have only applied our method to classification problems.
To go beyond, in the future, the method needs to be applied on deeper models such as Resnet \cite{He20152} and bigger datasets. Autoencoders, fully convolutional networks and segmentation problems are also an important focus. Furthermore, it would be interesting to generalize the $l_1/l_2$-norm to the $l_1/l_q$-norm to study its properties in more detail and improve on different model structures and learning problems.





%
\bibliographystyle{IEEEtran} 


\end{document}